# New Ideas for Brain Modelling 2

Kieran Greer, Distributed Computing Systems, Belfast, UK.
http://distributedcomputingsystems.co.uk
Version 1.1

*Abstract* - This paper describes a relatively simple way of allowing a brain model to self-organise its concept patterns through nested structures. For a simulation, time reduction is helpful and it would be able to show how patterns may form and then fire in sequence, as part of a search or thought process. It uses a very simple equation to show how the inhibitors in particular, can switch off certain areas, to allow other areas to become the prominent ones and thereby define the current brain state. This allows for a small amount of control over what appears to be a chaotic structure inside of the brain. It is attractive because it is still mostly mechanical and therefore can be added as an automatic process, or the modelling of that. The paper also describes how the nested pattern structure can be used as a basic counting mechanism. Another mathematical conclusion provides a basis for maintaining memory or concept patterns. The self-organisation can space itself through automatic processes. This might allow new neurons to be added in a more even manner and could help to maintain the concept integrity. The process might also help with finding memory structures afterwards. This extended version[1] integrates further with the existing cognitive model and provides some new conclusions.

## 1  Introduction

This paper describes a relatively simple way of allowing a brain model to self-organise its concept patterns through nested structures. For a simulation, time reduction is helpful and it would be able to show how patterns may form and then fire in sequence, as part of a search or thought process. The equation that is tested is possibly a simplified version of ex-isting ones ([13] equation 1, for example) which would also consider the synaptic connections in detail. This model is very much generalised and considers the pattern firing properties only. As such, it would be a quicker algorithm to use and so it may allow for more economic test runs that are concerned with this aspect in particular. The algorithm is attractive because it is still mostly mechanical and therefore can be added as an automatic process to any simulator. The process can also be used as a basic scheduling or counting mechanism and so it might be possible to add more mathematical operations to a brain-like model, without changing the basic components too much.

---

[1] This is an extended version of the paper accepted by The Science and Information Conference (SAI'14), London, 27 – 29 August, 2014.

The pattern model that is suggested is based on one of the most commonly occurring structures in the real world and so it is clearly understood and could be added to a computer program relatively easily. It is however a generalised high-level idea, without exact cell workings or synapse connections, for example. After the model and the reasons for suggesting it are described, some tests based on a relatively simple equation will be presented, to show the correctness of the idea. It will then be shown how the ensemble patterns can fit in with the current cognitive model and some other new ideas will be presented, along the lines of self-organisation and regulation.

The rest of this paper is organised as follows: section 2 describes some related work. Section 3 describes the main ideas of the new model. Section 4 describes some tests and results that confirm the main idea. Section 5 integrates the ensembles with the current cognitive model, while section 6 gives some conclusions on the work.

## 2    Related Work

This section is more biologically-oriented, where the author is not particularly expert, but the papers might make some relevant points. The aim is to show that the proposed structure is at least practical. The paper [13] is quite closely related and includes a number of important statements. It gives one example of an equation for the firing rate that includes the whole range of inputs, including external sensory and other neurons' excitatory/inhibitory input. It states that for the firing to be sustained, that is, to be relevant, re-quires sufficient feedback from the firing neurons, to maintain the level of excitation. Once the process starts however, this can then excite and bring in other neurons, when inhibitory inputs also need to fire, to stop the process from saturating. A weighted equation is given to describe how the process can self-stabilise if 'enough' inhibitory inputs fire and a comparison with the equation is given in section 4. The paper [10] also studies the real biological brain and in particular, the chemospecific steering and aligning process for synaptic connections. It notes that there are different types of neuron, synaptic growth and also theories about the processes. While current theory suggests that growth is driven by the neuron itself, that paper would require it to be driven almost completely by the charged 'signal'. Current theory also suggests that the neuron is required first, before the synapses can grow to it. However, they do note a pairwise chemospecific signalling process, as opposed to something that is just random and they also note that their result is consistent with the known preferences of different types of 'interneurons' to form synapses on specific domains of nearby neurons.

The paper [14] also describes how neurons can change states and start firing at different rates. The paper [11] describes that there are both positive and negative regulators. The positive regulators can give rise to the growth of new synaptic connections and this can also form memories. There are also memory suppressors, to ensure that only salient features are learned. Long-term memory endures by virtue of the growth of new synaptic connections, a structural change that parallels the duration of the behavioural memory. As the memory fades, the connections retract over time. So, there appears to be constructive synaptic processes and these can form memory structures. The paper [12] is more computer-based and describes tests that show how varying the refractory (neuron dynamics) time with relation to link time delays (signal) can vary the transition states. They note that it is required to only change the properties of a small number of driver nodes, which have more input connections than others. These nodes can control synchronization locally and they note that depending on the time scale of the nodes, some links are dynamically pruned, leading to a new effective topology with altered synchronization patterns. The structures tested are larger control loops, but it is interesting that the tests use very definite circular pattern shapes.

The work of Santiago Ramón y Cajal[2] has been suggested as relevant, in particular, with relation to pacemaker cells. This is definitely interesting and will be discussed in later sections. However, while Cajal appears to classify neurons, based on location defining their function, this paper does not consider different neuron types. It is only interested in location for allowing them to operate as part of a thinking process.

The author has proposed neural network or cognitive models previously [4][5][6][7] and it is hoped that this paper does not contradict that work. The aim has been to build a computer model that copies the brain processes as closely as possible, so as to realise a better or more realistic AI model. In particular, this paper might be considered an extension of [4]. It is still a computer program however and a close inspection of how the biological components work has only tentatively been introduced. The goal is to try and make the underlying processes as mechanical, or automatic, as possible, so that the minimum amount of additional intelligence is required for them to work. Earlier themes included dynamic or more chaotic linking, time-based events, pattern formation with state changes, clustering and even hierarchical structures with terminating states or nodes. This paper is associated with some of the earlier work, including the more chaotic neural network structures [4][5], or the pattern forming levels of the cognitive model [6][7].

## 3 Reasons For the Firing Patterns Model

It is important to remember that an energy supply is required to cause the neurons to fire. It is probably correct to think that the brain must receive a constant supply of energy to work. If a neuron fires, this would necessarily use up some of the energy, which is why the supply must be refreshed. If thinking about the single neuron, it is thought that ion channels cause the neuron activation, where pressure or force is not the main mechanism[3]. A neuron itself does not have the intelligence to fire, in the sense that it is reactive and not proactive. The fact that inhibitors are used to suppress the firing rate shows that the neurons cannot decide this for themselves. They also need an automatic mechanism to switch off. The activa-tion might be traced back to the external stimulus, which is a continuous energy source, although pressure would be another one [15]. Note also that the brain would be expected to give feedback, which in turn might change the input, and so on. Therefore, if considering the energy used by the system, it would make sense to nest sub-concepts, based on the idea of distance alone.

---

[2] http://www.scholarpedia.org/article/Santiago_Ramón_y_Cajal.
[3] Pressure is not very relevant for this paper, but was used as part of an earlier argument [4] to help the synaptic structures to grow and re-balance.

### 3.1 Sub-Concepts

When thinking about brain firing patterns, it appears to be very random and complex. Pictures or scans of activity however usually show distinct brain areas that are active, where this in itself is interesting. If the firing activity was completely random, then specific areas should possibly not be present, as synapses would travel in any direction to other neurons. So there is an indication here that the firing activity is contained. This then means that it could be inwards, or inside the originally activated area. This can sometimes be almost the whole brain, however. A simple example might also illustrate something. Thinking about a coffee cup, the cup itself can be imagined, sitting by itself. To expand the scenario, possibly a kettle is imagined, filling the cup with water. A specific action would also be invoked here. But is it the case that when the kettle is imagined, the whole kitchen scenario is retrieved? Even if this is done sub-consciously, it would put the scenario into a familiar context, when the kettle and subsequent actions are then easy. This is therefore interesting, because it suggests that the larger activity areas could be these larger concepts that then contain lots of smaller ones inside of them. The kitchen can contain a kettle and a cup, for example. If firing is restricted to the kitchen scenario, it is easy to imagine that it might activate the other smaller kitchen-related concepts. This is also a part of how we try to model the real world, mathematically or formally, in our processes or diagrams, for example.

Further, a single concept can be imagined by itself and even without a background. But the addition of context, invoked by an action or other object, forces the relevant background, even if it is relatively weak. So is the coffee cup and kettle driving the activation and triggering the kitchen that lives somewhere else, or is the span or area of activation now wide enough to activate part of the parent kitchen concept? If the firing was always inwards then activating a larger concept would be difficult, so at least some lateral or outwards positive activation is required. But then again, a coffee cup or kettle might be terminal states that are accessed directly (see possibly the θ value of equation 1 in [13]). As separate pattern groups also need to link, lateral signals could excite a general area between them as well. An action might even originate in a different brain region, bringing all of the connected areas into play. Figure 1 is a schematic of the general idea. There is a larger pattern with nested ones and some excitatory and some inhibitory signals. Traversing the larger area would bring in more of the background patterns or images. So the currently firing pattern is what defines the brain state. If there is no other way of controlling this, the ability to switch off the other areas in an automatic manner is required. If the parent provider encapsulates the new or most active next state, then this activity could be through a relatively simple and easy to understand process. The inhibitors will naturally send more negative feedback to their neighbouring environment, thereby weakening the parent signal compared to the new firing pattern. If new areas inside of them then become active, the process can repeat again. The most obvious catch is the fact that lateral linking and activation is always required and also from other brain regions that perform other functions. It is however, still a natural way to self-organise.

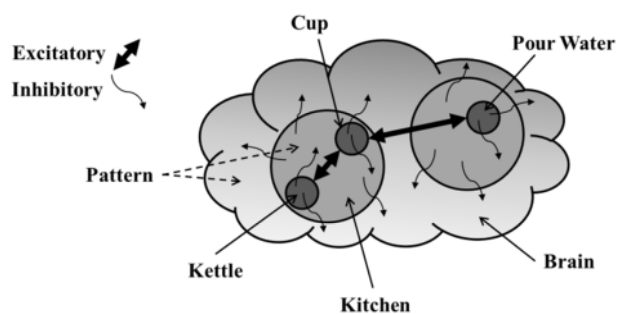

Figure 1. Example of Nested Concepts in a Brain Area.

## 3.2 Mathematical Processes

Another interesting use of the nested patterns is not to retrieve sub-concepts, but to implement a basic timer, counter or even battery, that could be part of more mathematical processes. The idea of battery, counter or timer here, refers to controlling the energy supply of a particular group of neurons. A more general supply is converted into one that can schedule something, or run for a pre-determined amount of time. It might be part of a whole cycle of pattern activations as follows: Some pattern activates another pattern that is the on-switch to a timer or battery. The on-switch activates the outer-most pattern of the nested group that makes up the new structure. This cycles inwards as described, until the inner-most pattern is activated. This might be 3 nested patterns, for example. Each nested pattern, when activated, might send a signal somewhere, but the inner-most one also sends a signal to the off-switch pattern that is beside the on-switch. The off-switch sends inhibitors to the on-switch, asking it to turn off. This then removes the signal inducer to the new structure and the whole cycle can stop.

As this is only an idea, an alternative and possibly better mechanism would be to slowly increase both the excitatory and inhibitory signals. The first activation phase from the outer-most pattern to the first nested one might not activate all neurons one level in. This also means however that they would not all send inhibitory signals back. So the outer-most pattern might be able to send several phases of signal before it receives an overwhelming inhibitory signal. The same situation can occur between any of the nested pattern sets. Continual activation signals can switch on more neurons the next level in, but then they also send more inhibitory signals back. If the excitatory signal is mostly inwards and the inhibitory one mostly outwards, this should result in the whole region eventually switching itself off. Slightly more doubtful: if the inhibitory signal only affects active neurons, then they can possibly fire in any direction, because the inner patterns will receive less than the outer ones, based on time events and so the outer ones will switch off first. Signal strength might also be a factor, if the excitatory signal is stronger than the inhibitory one. In that case, the inhibitory signal sent forwards to the nested patterns might not be strong enough to deter their activation; but the inhibitory signal sent back to the parent would then need to be from multiple nested or enclosed patterns, not just the immediate one. If signal strength is a factor, this would actually make the equation of section 4.1 even better and might also work better within a hierarchy setting, where signal is sent back to any 'upper' level, not just an 'enclosing' level. Also implicit then is the fact that there is a difference in how the excitatory and the inhibitory signals are created, where the excitatory ones have more direction.

So there would still be the desired and gradual build-up of signal and shut-down afterwards, even if the inter-relations between the basic components are slightly different. Note that these cases are started by a constant, external energy supply, which then gets shut-down or ignored. It would also be helpful if inhibitors could change a neuron state without switching it off completely and ideas of localised firing already exist [12]. The schematic of Figure 2 tries to describe the most general case. Some area of the brain excites and starts the outer-most pattern firing. This is the 'on' switch with a signal to the outer-most circle. The pattern cycles through to the inner-most one that can then ask for the provider to switch off. This is the 'off' switch. Each nested pattern can also send a signal somewhere else, which would implement the counting mechanism. The paper [13] notes another model that already exists called Synfire chains. Synfire chains fire in a definite outwards direction and offer some degree of control through firing stages at different levels. This then leads to problems of explosion from a sustained input and requires noise or other to control the firing rates. So the main question for the model of this paper is whether it can actually occur naturally, as other formations appear to be outward facing. It is worth remembering that clear pattern boundaries get created however and an inward firing process might be replaced by a hierarchical one, without changing anything else. See section 5.2.

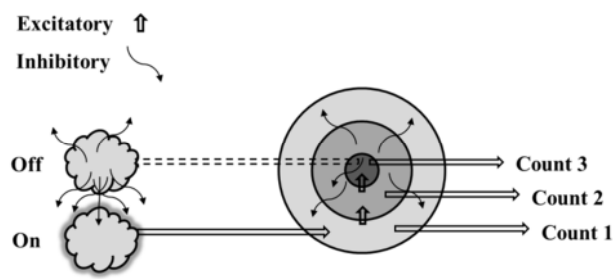

Figure 2. Example of a Timer, Counter or Battery.

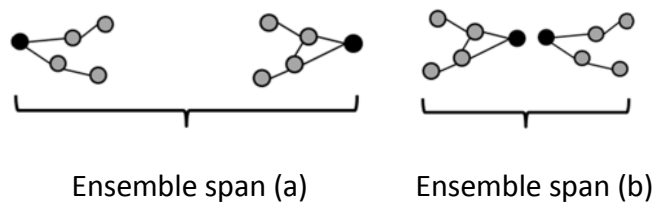

Ensemble span (a)     Ensemble span (b)

Figure 3. The two search process that ends with closer terminal states (b) is more economic.

As well as deciding to fire, this paper would require the neuron to intelligently control direction. Why would the neurons prefer to fire inwards instead of any direction? The theory of this paper however allows that intelligence to be replaced with an economic reason, based on the conservation of energy. If thinking about stigmergic systems, [2] for example, the ant colony selects the most economic path unintentionally and neurons equally influence each other. The idea of grouping more closely, neurons that fire at the same time, is also the well-known doctrine of Hebb [9]. The search process would also conceivably converge on terminal states [4], where Figure 3 could help to describe the economic argument. The idea of 'neurons that fire together wire together' requires a link between the two or more groups involved. If search occurs from a broad group to a smaller terminal state; then if that search is outwards, as in Figure 3a, the distance between the terminal states and the nodes in general is greater than if it is inwards, as in Figure 3b. Note in particular the case where the terminal states join to complete a circuit. Also, exactly as with stigmergy, if both pattern sets receive the same amount of energy, Figure 3b will reinforce more, because the signal can take a shorter route. That might just provide a reason why it is easier for the inward pointing search to then connect with another related search area, than the outward pointing one. Therefore, even by chance, a random process might prefer the inward facing groups.

## 4    Testing and Results

Testing of the theory can be carried out by implementing some basic reinforcement algorithms and updating specific node values, to simulate the different timings of the node pattern activations. The traditional increment/decrement reinforcement algorithm worked well enough to give the desired result. With that, the node value is incremented with excitatory input and decremented with inhibitory input. The decrement value can be weighted to be only a fraction of the increment. Some assumptions are made with regards to the neurons, which helps to simplify the problem further:

- Each neuron has only one excitatory output and one inhibitory output.

- The excitatory output goes only to the other neurons firing in the same pattern.
- The inhibitory output goes only to any other neuron that is in any parent pattern.
- Neurons are in the same pattern if they fire at the same time. This is measured by time increments $t_1 \ldots t_n$.

### 4.1 Test Conditions

The equation 1 for firing rate networks in [13] is probably a complete version of the equation that might be used. These tests only consider the excitatory/inhibitory part, to measure how the patterns will switch on and off through their interactions. The firing interactions are further restricted by the aforementioned assumptions. The resulting test equation for this paper could therefore be written as follows:

$$X_{it} = \sum_{p=1}^{P_i} Ept - \left(\sum_{k=P_j}^{l} \sum_{y=1}^{m} \sum_{j=1}^{n} (Hjy * \delta)\right)$$

where $y \neq t$ and $i \in P_i$ and $not\ j \in P_i$, and

$X_{it}$ = total input signal for neuron i at time t.

$E_p$ = total excitatory input signal for neuron p in pattern P.

$H_{jy}$ = total inhibitory input signal for neuron j at time y.

$\delta$ = weights inhibitory signal.

t = time interval for firing neuron.

y = time interval for any other neuron.

n = total number of neurons.

m = total number of time intervals.

l = total number of active patterns.

$P_i$ = pattern for neuron i.

P = total number of patterns.

In words, the tests measure how the total signal input to each neuron pattern changes. All neurons in the same pattern fire at the same time and send each other their positive signal. Any active neuron also receives a negative signal from any other nested pattern neuron. If two nested patterns are active, for example, the inner-most sends inhibitory signals to both the outer-most pattern and the first nested one. The first nested pattern sends inhibitory signals to just the outer-most one. Over time, neurons continue to fire based on - total pattern firing strength minus total inhibitory firing strength from all other nested patterns.

### 4.2 Test Results

The test results are quite straightforward and show the desired set of relative counts or signal strengths. Just the traditional increment/decrement algorithm is shown in Table **1**. There are 25 neurons in total and 5 in each nested pattern. The inhibitory signal is set to be half that of an excitatory one, but if a pattern only contains 5 neurons, that leaves a possible 20 other neurons that might send inhibitory signals. Each firing cycle activates a new nested pattern, until all patterns are active. After that, each firing cycle would update signals from all patterns. The inhibitory signal is sent from the inner pattern to its outer ones only, so the inner-most one does not receive inhibitory signals.

Table 1. Relative Pattern Strengths after Firing Sequences.

| Neurons | t = 3 | t = 4 | t = 5 |
|---|---|---|---|
| 1 | 7.5 | 5.0 | 0.0 |
| 2 | 7.5 | 5.0 | 0.0 |
| 3 | 7.5 | 5.0 | 0.0 |
| 4 | 7.5 | 5.0 | 0.0 |
| 5 | 7.5 | 5.0 | 0.0 |
| 6 | 7.5 | 7.5 | 5.0 |
| 7 | 7.5 | 7.5 | 5.0 |
| 8 | 7.5 | 7.5 | 5.0 |
| 9 | 7.5 | 7.5 | 5.0 |
| 10 | 7.5 | 7.5 | 5.0 |
| 11 | 5.0 | 7.5 | 7.5 |
| 12 | 5.0 | 7.5 | 7.5 |
| 13 | 5.0 | 7.5 | 7.5 |
| 14 | 5.0 | 7.5 | 7.5 |
| 15 | 5.0 | 7.5 | 7.5 |
| 16 | 0.0 | 5.0 | 7.5 |
| 17 | 0.0 | 5.0 | 7.5 |
| 18 | 0.0 | 5.0 | 7.5 |
| 19 | 0.0 | 5.0 | 7.5 |
| 20 | 0.0 | 5.0 | 7.5 |
| 21 | 0.0 | 0.0 | 5.0 |
| 22 | 0.0 | 0.0 | 5.0 |
| 23 | 0.0 | 0.0 | 5.0 |
| 24 | 0.0 | 0.0 | 5.0 |
| 25 | 0.0 | 0.0 | 5.0 |

When all patterns are active, the inhibitory signal builds up to overwhelm the excitatory signal. This of course, depends on the pre-set relative strengths and numbers of excitatory and inhibitory signals. Neurons 1 to 5 are the outer-most pattern. Neurons 6 to 10 are the first nested pattern and so on, until neurons 21 to 25 are the inner-most nested pattern. At time t1, the first pattern only fires (neurons 1 to 5). At time t2 pattern 1, then pattern 2 fires. At time t3, pattern 1, then pattern 2, then pattern3 fire, and so on. The outer patterns have more excitatory input to start with, but as the other patterns switch on and send negative feedback, eventually they will switch off the outer patterns. This would then actually starve the inner patterns of input, until they switch off as well.

## 5 Cognitive Model

It turns out that the nested ensembles can fit into the current cognitive model, described in [5][6][7] and most recently in [4]. All of the elements mentioned in earlier papers are still relevant and so the model can be refined further. While it may not be 100% correct biologically, it is becoming quite detailed and still consistent over the main ideas.

### 5.1 Hierarchical Nesting

The first thing to think about again is the regions, or nested regions, of pattern ensembles. As the individual elements are likely to be located randomly, duplication can help. For example, if the concepts in question are duplicated they can occur in different locations and collections, but the permanent ensemble will probably be formed where they are located closest to each other. It is noted that duplication also occurs because different parts of the brain store the same concept for different reasons. The most economic group might complete the connections first, satisfy the input requirements and reinforce the most, as in stigmergy. Therefore, duplication makes it more likely that any ensemble can form, or if the neuron is missing, does it just get created? It is also noted that neurons are created from some sort of chemical reaction and are not required to grow at the end of a synapse, or anything like that. So the stimulus itself can create new neurons as needed.

If looking at the neural network model of [7] again - trying to justify everything is silly, but a similar situation that favours a unique set of closer grouped entities might be relevant. It was also shown in the neural network that noisy input could be filtered out more easily, which might also be a helpful feature for the nested ensembles. The noise would be filtered out more easily because it might not be consistently the same in each group, whereas each specific concept would be. Keeping the individual groups separate does not allow random noise to form into more common clusters. The ensembles are then connected through the hierarchy. It has also been suggested that the physical space and the logical space are different. For a comparison with the neural network, the hidden layer(s) is a combination of the nodes in the level below. For a nested ensemble, this is simply the parent or enclosing region of the group in question. This can continue up to the outer-most region. That would be the largest region, but would represent the most global and general concept as well. So the hierarchy is from the smaller nested regions to the larger enclosing one. The idea of a trigger, as shown in the earlier figures [4][6], is also appropriate and is even represented in Figure 1 of this paper. It could be the lateral connecting synapse between the two inner circles that might be used to link-up different types of concept into logical sequences.

Also, can Figure 3 of this paper be compared to the concept trees of other papers [4], figure 6, for example? It is shown again here in Figure 4. Static knowledge also needs to be learned and base nodes at the bottom of trees might provide activation paths to the groups or concepts at the leaves that then get arranged further through time. The dynamic time-based layer is maybe where neurons groups are initially connected-up to form ensembles, but it is better to have 3 concept groups there instead of 1 and have them link-up in another level if required. Maybe that way, the ensemble can still be a more chaotic arrangement, while the structured process that builds the concept trees remains as well. So the ensembles are also somewhere along this first time-based line and then up through the whole hierarchy.

Figure 3 of this paper maybe has the node structures 180 degrees the wrong way round, where the broader regions should try to connect with each other. The idea of a construction process in one direction and a search in the opposite direction is also again consistent. We can guess that the construction process for the ensemble hierarchy is again from the static concept trees to the dynamic global concepts. If that is the case, we would in fact see small details first and then aggregate them into larger entities. This is good for another reason. There is then a direct path to these smaller concepts that gets added during their formation and would allow them to be accessed directly, as in memory retrieval. We then perform a search in the opposite direction, from our general impression to the finer details. We find the general region first and then search 'inside' of that for specific information. Or maybe if the search is unclear, a larger area must be activated first, as in browsing. If these end with concept tree-like structures, then the small ensembles can even trace down to the single terminating base nodes that may allow a common connection with other brain regions.

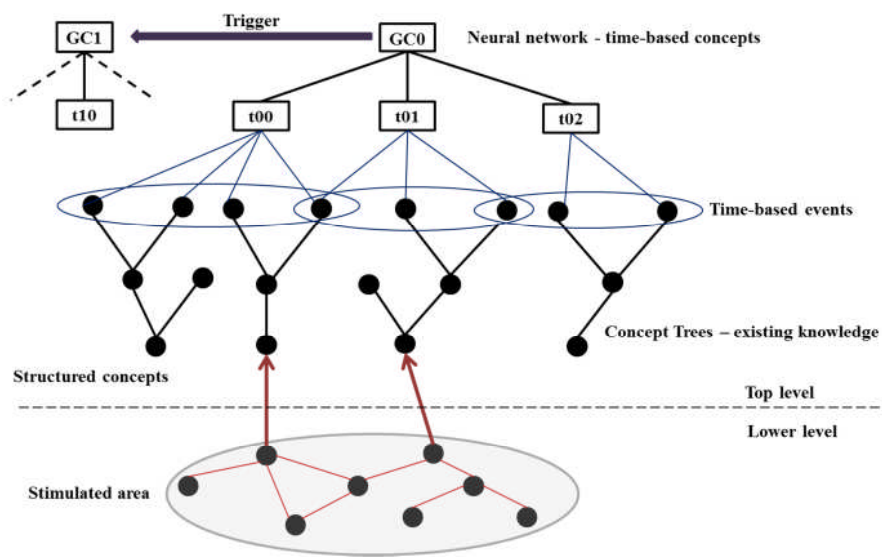

Figure 4. Integration of elements into the Cognitive Model, also from [4]

Looking at the actual human brain, the search probably starts in the neocortex [8] which is the thin upper layer that envelops most of the rest of the brain. So that is OK and is a top-down cognitive search. This means that the initial learning process must be more bottom-up and possibly carried out more through observation than prediction or interpretation. But that is probably OK as well. The hippocampus is supposed to be where memories are stored, or at least, is critical to their formation. It is a separate structure to the neocortex, as the concept trees are to the ensemble hierarchy. Synfire chains [13] demonstrate a cascading activation process over a hierarchical structure that could represent a search process over the ensemble hierarchy. This has been pointed out earlier in section 3.2 as opposite to the inward firing process assumed by the nested ensembles, but it may in fact be the same. The inward firing of this paper is from the outer region to the nested regions, which is the same as from the upper hierarchical level to the multiple lower levels. It is just the physical representation that was confusing. One difference might be that while the outer-most region represents one basic concept, it might contain more neurons numerically. These then excite more concepts in the next level but they are each represented by smaller numbers of neurons. This is interesting in itself, if a neuron must represent something specific, or can it be purely functional? Are some neurons used simply to activate the next level upon request? It might be worth mentioning the pyramidal neurons found by Cajal, for section 5.2 in particular. These have multiple inputs at the base and an output axon with synapses that span any area and so can flow in any direction. They would also be ideal as the base neurons of (concept tree) memory structures.

## 5.2   Circuit Reinforcement and Balancing

It has been shown previously [2][3] how ant or termite colonies can collectively determine an optimal route through a basic reinforcement mechanism, without any prior or global knowledge about the route. Each insect leaves a trace that is read by the other insects and it is only that process which determines the optimal route. The Figure 3 of his paper shows how a similar process will encourage the closer sets of neurons to form together first, as the overall distances are less. This is also based on local information only and with no prior knowledge. As these insects are believed to work through a nervous system and not a brain as humans have, it is reasonable to apply a similar type of process for the pattern constructions. Research has already shown how firing neuron activity can saturate ([13], for example), but this does not mean a free energy supply. It must also be considered that if a signal is sent from a neuron to more than one place, then it can only be at a fraction of the single signal strength and so it might require a faster firing rate to maintain the same strength to multiple places. Inhibitors probably help to self-regulate the firing rate. The more that feeder neurons activate an area, the more it will send out inhibitors that in turn might slow down the feeder activation signals, until some happy medium is met.

Distance is also important along a single route and must be considered along with the energy consumption and the neuron threshold value. More energy would be required to force a signal along a longer path, where repeated firing by the feeder neurons would probably be required to maintain the signal flow. The paper [4] includes very basic equations that consider disruption of the signal over some distance. There does not have to be a forceful disruption for this aspect, only the natural impedances, but similar types of equation can be applied. These will be described in quite abstract and general terms, so it is the idea of them and not exact values that is important. For a signal to be maintained therefore, we need to consider how much energy might be lost over a particular distance. For example:

- Let $T_m$ be the threshold for the neuron that is to be activated.
- Let $I_s$ be the input signal from the feeder neuron. Consider a single line or path, with just one feeder neuron to the next neuron.
- Let $d$ be the distance from the feeder neuron to the activation neuron.
- Let $\alpha$ be the amount of signal that is lost per unit distance.

Then, even if a neuron can eject the same amount of output that is received as input, the output signal required by the feeder neuron might be:

$I_{sn} + (d \times \alpha)$.

Or an excess of *(d × α)* is required to cover the distance of the signal to the next neuron.

As a neuron can act as a capacitor, this can mean that multiple signals are sent and stored, until the cumulative result fires the activation neuron. If the activation neuron is then required to activate another neuron further along, it faces the same problem. Thinking again in very abstract terms, the additional required signal becomes a multiple of the requirements of each individual neuron along the path. For example, if neuron 1 needs to fire twice to send enough signal to activate neuron 2 and neuron 2 needs to fire twice to send enough signal to activate neuron 3; then neuron 1 needs to fire 4 times to allow neuron 3 to fire. As far as balancing the neurons' organisation inside of any ensemble is concerned, this is actu-ally a good result. Consider a line of these neurons that span a particular region, where the distances between them varies as follows:

**N1**  5  **N2**  2  **N3**  2  **N4**  2  **N5**  10  **N6**  10  **N7**

When clustering, it can be a bit localised, where typically the closest distance between any two points is measured. If the neurons N1 to N7 represent the points along a line, then neuron N3 would typically be considered to be at the centre of a cluster. If a region spans the whole area from N1 to N7 however, we would prefer any new neurons that get added to be evenly spaced and not to amass around neuron N3, being related with the closest local distances. If we use a cumulative multiplication of the required signal amount across the whole line, then the centre is determined by the distance over the whole region only and not between individual neurons. In the example, if each numerical value is the additional required signal to reach the next neuron, then a value of 2 is required to go from N3 to N4, for example. But N4 then also needs 2 to go to N5, and so on. We are also trying to minimise the amount of energy that is required and also allow synapses to travel in both directions from a single neuron output. With a cumulative multiplicative count, a cluster centre like N3 would travel to N1 in one direction and N7 in the other, requires (2 x 5) + (2 x 2 x 10 x 10) = 410 additional firings. Neuron N5 looks like it is at the edge of a cluster, but to span the same distance, it requires (5 x 2 x 2 x 2) + (10 x 10) = 140 additional firings. The larger distances are prohibitive for neurons not located in the centre, distance-wise. So if this region was excited from the centre, which appears to be the most economic, it would prefer neuron N5 over neuron N3. This might also help to space the creation of new neurons better, because the energy or stimulus is always located in the centre and not necessarily in the densest region. There is also a robustness or integrity reason why a central activation is better. If the activation path to the ensemble was at the edge and not at the centre, then it might be easier to change the ensemble concept by adding neurons to the other side of the edge. If the activation path is to the centre, then even if something new gets added, the original concept can maintain its original meaning, where the change is an addition rather than a radical shift to something else.

## 5.3 One-way System?

With clear input and output sets to neurons and the requirements to complete circuits, it may be thought that most processes are one-way systems that are cyclic in nature. Although, recent brain models show fibrous or tree-like branching properties, at some level of granularity. Cyclic or circular completions are very easy to understand, but assume that the whole process is 'as one', where it might then be further assumed that the signals that flow need to be very similar. While they match with the known neuron functionality, there may be other problems. For one thing, there are definite regions in the brain and so signals would need to cross boundaries and even functionality type. Similarly, if the signal flows in one direction, for example, from the top of the hierarchy to the bottom in the neocortex (or the simulated neural network model), then to complete the cycle it needs to flow back up again, which is again possibly changing the functionality. It is noted that a cyclic completion can be more at the end of an activation chain than its full length and therefore another suggestion can also be made. An alternative way to complete a circuit is to have (at least) two halves that meet or join. This is attractive for a number of reasons. One is the distance reason again. It would be possible for any brain region to start sending a signal into the 'middle area' of the brain, for example and have it travel anywhere else. It might then meet a matching signal from the receiving area. The two can join to reinforce and complete a slightly different mode of circuit. This could even be at the other side of the brain and so there does not have to be a long feedback loop to the original source. This also means that when a stimulus is set off, the neocortex or some other area can independently start working to satisfy it and different regions can work in parallel. It can run through its own structures to try to match the signal from other places. The signals that meet might then actually collide to join up, instead of flow in the same direction together, but they still complete the circuit and the interaction might even facilitate the essential ingredients of resonance [1][4][5], if the firing rates or signals strengths also match. The signals can maybe meet at the region boundaries, such as somewhere around the concept trees layer in the computer-based model.

It is still possible to build a similar system with the existing components that would incorporate the cyclic reinforcement more. The neurons can still have only one input and one out-put function, but some can face each other and then be joined by synapses that fuse with another set of neurons running sideways. The opposing signals induce the lateral activity that can flow through the sideways facing neurons and even re-connect to complete some type of circuit. The Figure 5 shows this schematically and might be compared with the ideas of pressure in [4].

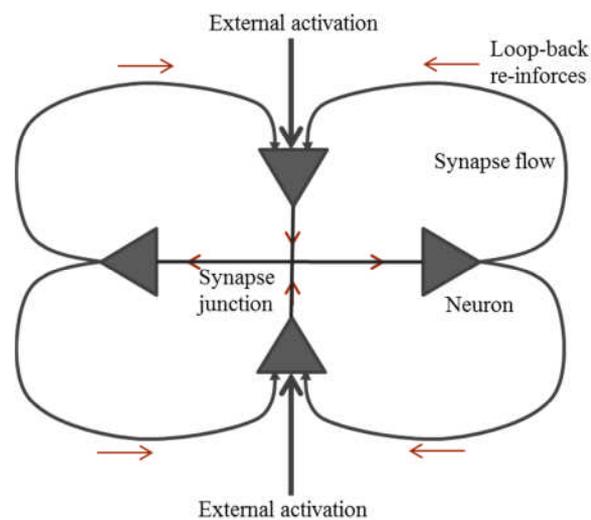

Figure 5. Top and bottom neurons join and activate lateral ones through synapse junctions, to register enhanced signals.

Imagine that the two brain regions are at the top and bottom of this figure. They send signals (external activation) from their own directions to excite the top and bottom neurons respectively. These have joins or links in the form of a synapse junction that includes sideways or lateral connections. The signal is forced through the lateral connections to activate the sideways facing neurons. They might even loop back to reinforce the signal, but only very locally, or in the specific region that represents the desired search result and neural ensemble. There could be less force if the top-bottom facing neurons actually joined at the lateral neurons instead of at the junction, but their creation or initial meeting might be from a straight connection between them, which is an easier automatic join. So the whole area, including the lateral connections, might grow in a normal manner, with new neurons or synapses being added to places that are more frequently used.

# 6  Conclusions

The purpose of this paper is to show how nested, or more specific patterns, may become the main focus in a generally excited area. They might even be used as part of more complex mathematical thought processes. Rather than the exact details of how they might be created, or link to each other, etc., the paper describes how they might be useful as a simplified design. Simulation would be easier if the interaction between the patterns only was considered, using a general equation for their relative strengths and ignore exact synaptic connections. The mechanical processes can work with a minimum of complexity and would allow these patterns to form and fire in sequence. It would also realise some level of natural order, which would be better than the very random and chaotic structures that appear to be present. It is interesting that a self-organising process might naturally prefer a nested structure, certainly to one that faces outwards.

The second purpose of this paper is to integrate the new findings into the whole cognitive model. It appears that the nested ensembles fit-in almost seamlessly with the existing ideas and through studying them, other helpful information has been obtained. In particular, there are several examples of how the processes can naturally regulate themselves and perform the type of functionality that you might think requires some level of intelligence. With respect to automatic processes, the nesting allows for the idea of terminal or end states, which can help with search processes. The act of searching into a smaller region as opposed to a larger region might also make the search process easier. Even just the signal strength can help with managing pattern transitions as part of an automatic process, so there is quite a lot that can be achieved with the basic components that are known about. As described in section 5, the mechanism is still compatible with earlier work.

The self-organisation can also space itself through automatic processes. This might allow new neurons to be added in a more even manner and could help to maintain the concept integrity. A link at creation-time, to the centre of the concept can also help with finding it directly, possibly as part of a memory structure. The idea of resonance being important is also enhanced, if some form of joining is to be preferred over the less violent reinforcement through cyclic links, or complementary with it. But then, each brain area can keep its own functionality and have a sort of interface. Also, the earlier ideas of the dynamic hierarchical network joining with the static knowledge-based one is still central to the whole architecture and even small pieces of evidence from the real biological world can help to support the ideas, where established theories are not so clear.

## References

[1] Carpenter, G., Grossberg, S., and Rosen, D. (1991). Fuzzy ART: Fast stable learning and categorization of analog patterns by an adaptive resonance system. Neural Networks, Vol. 4, pp. 759–771.

[2] Dorigo, M., Bonabeau, E. And Theraulaz, G. (2000). Ant algorithms and stigmergy, Future Generation Computer Systems, Vol. 16, pp. 851 – 871.

[3] Grassé P.P. (1959). La reconstruction dun id et les coordinations internidividuelles chez Bellicositermes natalensis et Cubitermes sp., La théorie de la stigmergie: essais d'interprétation du comportement des termites constructeurs, Insectes Sociaux, Vol. 6, pp. 41-84.

[4] Greer, K. (2014). New Ideas for Brain Modelling, accepted for publication by Journal of Information Technology Research (JITR), Special issue on: From Natural Computing to Self-organizing Intelligent Complex Systems, Eds. Barna László Iantovics, Constantin-Bala Zamfirescu, Kenneth Revett and Adrian Gligor, IGI Global, published on arXiv at http://arxiv.org/abs/1403.1080.

[5] Greer, K. (2013). Artificial Neuron Modelling Based on Wave Shape, BRAIN. Broad Research in Artificial Intelligence and Neuroscience, Vol. 4, Issues 1-4, pp. 20- 25, ISSN 2067-3957 (online), ISSN 2068 - 0473 (print).

[6] Greer, K. (2012). Turing: Then, Now and Still Key, book chapter in: 'Artificial Intelligence, Evolutionary Computation and Metaheuristics (AIECM) - Turing 2012', Eds. X-S. Yang, Studies in Computational Intelligence, 2013, Vol. 427/2013, pp. 43-62, DOI: 10.1007/978-3-642-29694-9_3, Springer-Verlag Berlin Heidelberg.

[7] Greer, K. (2011). Symbolic Neural Networks for Clustering Higher-Level Concepts, NAUN International Journal of Computers, Issue 3, Vol. 5, pp. 378 – 386, extended version of the WSEAS/EUROPMENT International Conference on Computers and Computing (ICCC'11).

[8] Hawkins, J. and Blakeslee, S. On Intelligence. Times Books, 2004.

[9] Hebb, D.O. (1949). The Organisation of Behaviour.

[10] Hill, S.L., Wang, Y., Riachi, I., Schürmann, F. and Markram, H. (2012). Statistical connectivity provides a sufficient foundation for specific functional connectivity in neocortical neural microcircuits, Proceedings of the National Academy of Sciences.

[11] Kandel, E.R. (2001). The Molecular Biology of Memory Storage: A Dialogue Between Genes and Synapses, Science magazine, Vol. 294, No. 5544, pp. 1030 - 1038.

[12] Rosin, D.P., Rontani, D., Gauthier, D.J. and Scholl, E. (2013). Control of synchronization patterns in neural-like Boolean networks, arXiv preprint repository, http://arxiv.org.

[13] Vogels, T.P., Kanaka Rajan, K. and Abbott, L.F. (2005). Neural Network Dynamics, Annu. Rev. Neurosci., Vol. 28, pp. 357 - 376.

[14] Waxman, S.G. (2012). Sodium channels, the electrogenisome and the electrogenistat: lessons and questions from the clinic, The Journal of Physiology, pp. 2601 – 2612.

[15] Willie, C.K., Macleod, D.B., Shaw, A.D., Smith, K.J., Tzeng, Y.C., Eves, N.D., Ikeda, K, Graham, J., Lewis, N.C., Day, T.A. and Ainslie, P.N. (2012), Regional brain blood flow in man during acute changes in arterial blood gases, The Journal of Physiology, Vol. 590, No. 14, pp. 3261–3275.